%% file: main.tex
\pdfoutput=1

\documentclass[11pt]{article}
\usepackage{arabtex}
\usepackage[]{EMNLP2023}
\usepackage{lipsum}
\usepackage{times}
\usepackage{latexsym}
\usepackage{arydshln}

\usepackage[T1]{fontenc}

\usepackage[utf8]{inputenc}
\usepackage{xstring}
\usepackage{cleveref}
\usepackage{arabtex}
\usepackage{utf8}
\usepackage{color}
\usepackage{arydshln}
\usepackage{adjustbox}
\usepackage{float}
\restylefloat{table}
\usepackage{array,booktabs,makecell}
\usepackage{multirow}
\usepackage{graphicx}
\usepackage{color, colortbl}
\usepackage{xstring}
\usepackage{microtype}
\usepackage{color,soul}

\usepackage{multirow}
\usepackage{graphicx}

\usepackage{listings}
\usepackage{color}

\definecolor{dkgreen}{rgb}{0,0.6,0}
\definecolor{gray}{rgb}{0.5,0.5,0.5}
\definecolor{mauve}{rgb}{0.58,0,0.82}

\usepackage{inconsolata}
\usepackage{arydshln}
\newcolumntype{H}{>{\setbox0=\hbox\bgroup}c<{\egroup}@{}}
\usepackage{cancel}
\usepackage{ulem,xpatch}

\usepackage{hyperref}

\definecolor{blush}{rgb}{0.87, 0.36, 0.51}

\definecolor{dgreen}{rgb}{0.0, 0.5, 0.0}

\usepackage{enumitem}

\usepackage{tikz}
\def\checkmark{\tikz\fill[scale=0.5](0,.35) -- (.25,0) -- (1,.7) -- (.25,.15) -- cycle;} 
\title{NADI 2023:\\The Fourth Nuanced Arabic Dialect Identification Shared Task}

\author{Muhammad Abdul-Mageed,$^\lambda$$^,$$^\xi$ AbdelRahim Elmadany,$^\lambda$ Chiyu Zhang,$^\lambda$ \\ \textbf{El Moatez Billah Nagoudi},$^\lambda$ \textbf{Houda Bouamor},$^\delta$ \textbf{Nizar Habash}$^\mu$\\
$^\lambda$The University of British Columbia, Vancouver, Canada; 
$^\xi$MBZUAI, Abu Dhabi, UAE\\
$^\delta$Carnegie Mellon University in Qatar, Qatar\\
$^\mu$New York University Abu Dhabi, UAE\\
  {\tt \{muhammad.mageed@,a.elmadany@,chiyuzh@mail,moatez.nagoudi@\}.ubc.ca} \\
  {\tt ~~ hbouamor@cmu.edu ~~~ nizar.habash@nyu.edu}\\
  }

\begin{document}
\setcode{utf8}
\setarab 
\maketitle

\begin{abstract}
We describe the findings of the fourth Nuanced Arabic Dialect Identification Shared Task (NADI 2023). The objective of NADI is to help advance state-of-the-art Arabic NLP by creating opportunities for teams of researchers to collaboratively compete under standardized conditions. It does so with a focus on Arabic dialects, offering novel datasets and defining subtasks that allow for meaningful comparisons between different approaches. NADI 2023 targeted both dialect identification (Subtask~1) and dialect-to-MSA machine translation (Subtask~2 and Subtask~3). A total of $58$ unique teams registered for the shared task, of whom $18$ teams have participated (with $76$ valid submissions during test phase). Among these, $16$ teams participated in Subtask~1, $5$ participated in Subtask~2, and $3$ participated in Subtask~3. The winning teams achieved  $87.27$
 F\textsubscript{1} on Subtask~1, $14.76$ Bleu in Subtask~2, and $21.10$ Bleu in Subtask~3, respectively. Results show that all three subtasks remain challenging, thereby motivating future work in this area. We describe the methods employed by the participating teams and briefly offer an outlook for NADI.

\end{abstract}
\input{intro}

\input{lit}
\input{tasks}
\input{datasets}
\input{results}

\input{conc}
\input{limitations_ethics}
\input{ack}
\normalem
\bibliography{dlnlp,NADI2023-paper,team_referances}
\bibliographystyle{acl_natbib}




\end{document}

%% file: intro.tex
\section{Introduction}\label{sec:intro}


\textit{Arabic} is a term usually used to collectively refer to a host of languages and language varieties, rather than a single language. While most of these languages and varieties are similar to one another, there can be significant differences between some of them. For example, Egyptian Arabic and Moroccan Arabic are not mutually intelligible. Arabic can also be classified into three broad categories, classical, modern standard, and dialectal. Of these, \textit{Classical Arabic (CA)} represents the variety used in old forms of literature such as poetry and the Qur'an, the Holy Book of Islam. Association with religion and literary expression endows CA with prestige, and it continues to be used to date side by side with other varieties. \textit{Modern Standard Arabic (MSA)}~\cite{badawi1973levels,mageed2020microdialect} is a modern-day variety that is more familiar to native speakers and is usually employed by pan-Arab media organizations, government, and in education. The third category, \textit{Dialectal Arabic (DA)}, is itself a superclass that is collectively assigned to a host of varieties that are sometimes defined regionally (e.g., Gulf, Levantine, Nile Basin, and North African~\cite{Habash:2010:introduction,mageed2015Dissert}), but are increasingly recognized at the more nuanced levels of country or even sub-country~\cite{Bouamor:2018:madar,mageed2020microdialect}). NLP treatment of Arabic dialects has thus far focused more on dialect identification~\cite{mageed2020microdialect,bouamor2019madar,Darwish:2018:multi-dialect}, machine translation (MT)~\cite{zbib2012machine}, morphosyntax~\cite{obeid-etal-2020-camel}. 
\begin{figure}[t]
\vspace{1.5cm}
  \begin{center}
  \frame{\includegraphics[width=\linewidth]{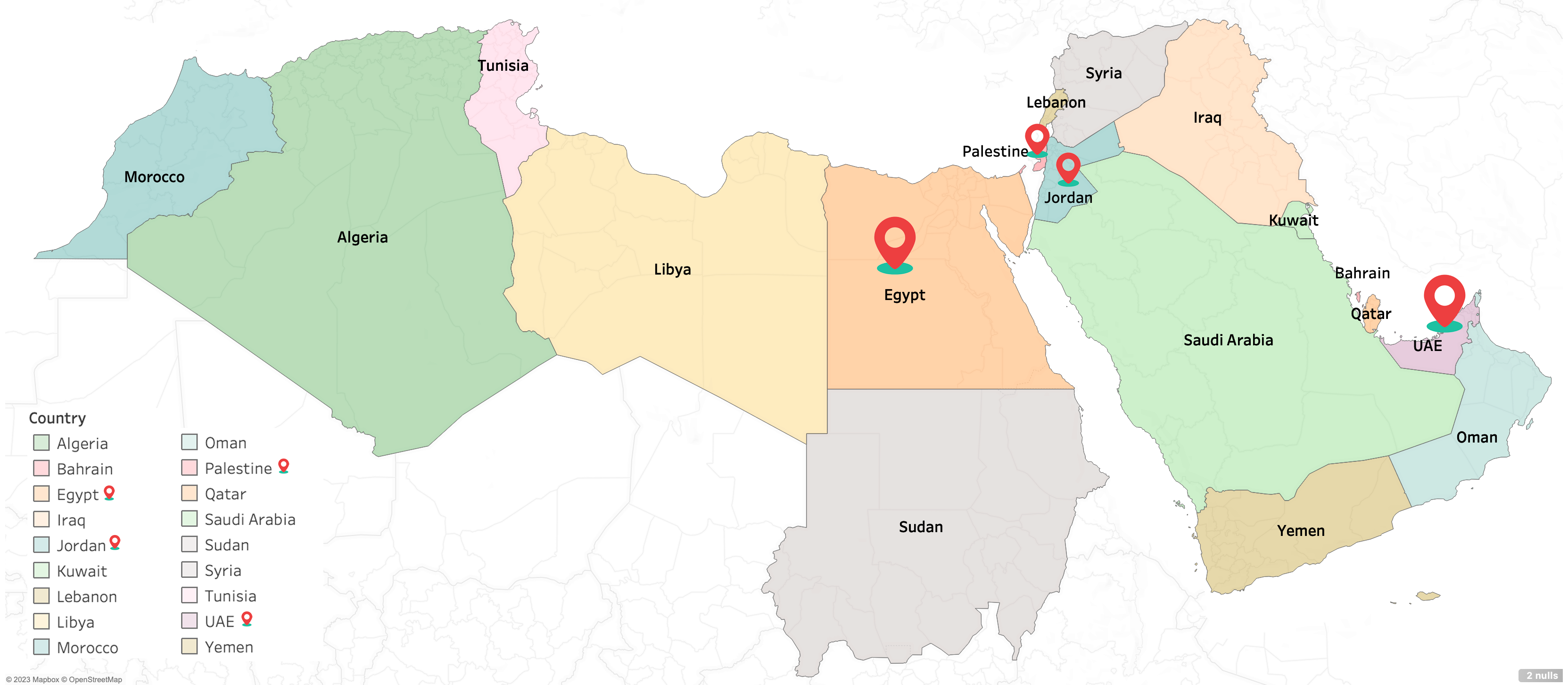}}
  \end{center}
\caption{A map of the Arab World showing the $18$ countries in the \textit{Subtask~1} dataset and the $4$ countries in the \textit{Subtask~2} and \textit{Subtask~3} datasets. Each country is coded in a color different from neighboring countries. Subtasks~2 and 3 countries are coded as red pins.}
\label{fig:map}
\end{figure}

\textit{Dialect identification} is the task of automatically detecting the source variety of a given text or speech segment, and is the main focus of the current work where we introduce the findings and results of the fourth Nuanced Arabic Dialect Identification Shared Task (NADI 2024). The main objective of NADI is to encourage research on Arabic dialect processing by offering datasets and facilitating diverse modeling opportunities under a common evaluation setup. The first instance of the shared task, NADI 2020~\cite{mageed:2020:nadi}, focused on province-level dialects. NADI 2021~\cite{mageed:2021:nadi}, the second iteration of NADI, focused on distinguishing both MSA and DA according to their geographical origin at the country level. The third instance, NADI 2022~\cite{mageed:2022:nadi}, investigated both Arabic dialect identification and dialectal sentiment analysis. NADI 2023, the current edition, continues this tradition of extending to tasks beyond dialect identification. Namely, we propose new subtasks focused at machine translation from Arabic dialects into MSA. 

More concretely, NADI 2023 shared task is comprised of three subtasks: \textbf{Subtask~1} on dialect identification, while  \textbf{Subtask~2} and \textbf{Subtask~3} are on dialect MT. The difference between Subtask~2 and Subtask~3 is that the former is a \textit{closed track} where participants are allowed to use only our provided training data, whereas the latter is \textit{open track} and so allows participants to train their systems on any additional datasets so long as these additional training datasets are public at the time of submission. While we invited participation in any of the three subtasks, we encouraged teams to submit systems to \textit{all} subtasks. By offering three subtasks, our hope was to receive systems that exploit different methods and architectures. Many of the submitted systems investigated diverse approaches, thus fulfilling our objective. A total of $58$ unique teams registered for NADI 2023. Of these, $18$ unique teams actually made submissions to our leaderboard (n=$76$ valid submissions during test phase). We received $14$ papers from $14$ teams, of which we accepted $13$ for publication. Results from participating teams show that both dialect identification at the country level and dialectal MT remain challenging even to complex neural methods. These findings clearly motivate future work on all tasks. 

The rest of the paper is organized as follows: Section~\ref{sec:lit} provides a brief overview of Arabic dialect identification and sentiment analysis. We describe the two subtasks and NADI 2023 restrictions in Section~\ref{sec:tasks}. Section~\ref{sec:dataset-eval} introduces shared task datasets and evaluation setup. We present participating teams and shared task results and provide
a high-level description of submitted systems in Section~\ref{sec:results}. We conclude in Section~\ref{sec:conc}.

%% file: lit.tex
\section{Literature Review}\label{sec:lit}

\subsection{Arabic Dialects} 
As stated earlier, Arabic can be broadly categorized into CA, DA, and MSA. While  CA and MSA have been examined extensively~\cite{Harrell:1962:short,Cowell:1964:reference,badawi1973levels,Brustad:2000:syntax,Holes:2004:modern}, DA became the center of attention only relatively recently. A significant challenge in studying DA has been the scarcity of resources. This prompted researchers to create new DA datasets, usually targeting a limited number of specific regions or countries~\cite{Gadalla:1997:callhome,diab2010colaba,al2012yadac,sadat2014automatic,Smaili:2014:building,Jarrar:2016:curras,Khalifa:2016:large,Al-Twairesh:2018:suar,alsarsour2018dart,kwaik2018shami,el-haj-2020-habibi}. This was followed by several works that introduced multi-dialectal datasets and models for region-level dialect identification~\cite{zaidan2011arabic,Elfardy:2014:aida,Bouamor:2014:multidialectal,Meftouh:2015:machine}.
The initial Arabic dialect identification shared tasks were part of the VarDial workshop series, primarily utilizing transcriptions of speech broadcasts~\cite{malmasi2016discriminating}. 
This was followed by creation of the Multi-Arabic Dialects Application and Resources project (MADAR), which provided finer-grained data and a lexicon~\cite{Bouamor:2018:madar}. Although MADAR's dataset was used for identifying dialects at both the country and city levels~\cite{Salameh:2018:fine-grained,obeid-etal-2019-adida}, the fact that it is commissioned, rather than naturally occurring, makes it not be optimal for dialect identification especially in contexts such as social media. 

Subsequently, larger datasets that cover between 10 to 21 countries were introduced~\cite{Mubarak:2014:using,Abdul-Mageed:2018:you,Zaghouani:2018:araptweet, abdelali-etal-2021-qadi,issa-etal-2021-country,baimukan2022hierarchical,althobaiti2022creation}. The majority of these datasets are compiled from social media posts, especially Twitter. 
Other works collect data at a more granular level. For instance,~\citet{mageed2020microdialect} introduces a Twitter dataset along with several models to identify variations in Arabic dialects at the country, province, and city levels.~\newcite{althobaiti2020automatic} provides an overview of computational work on Arabic dialects. More recently, benchmarks such as ORCA~\cite{elmadany-etal-2023-orca} and DOLPHIN~\cite{nagoudi2023dolphin} boast dialectal coverage. The NADI shared task continues to lead efforts on providing datasets and common evaluation settings for identifying Arabic dialects~\cite{mageed:2020:nadi,mageed:2021:nadi, mageed:2022:nadi}.  

\subsection{Machine Translation of Arabic Dialects}
Several studies focus on machine translation of Arabic dialects. For example,~\newcite{zbib2012machine} demonstrate effects of using both MSA and DA data on performance of Dialect/MSA to English MT.~\newcite{sajjad2013translating} employs MSA as an intermediary language for translating Arabic dialects into English.~\newcite{salloum2014sentence}  examine the impact of sentence-level dialect identification and various linguistic features on Dialect/MSA to English translation.~\newcite{guellil2017neural} propose a neural system for translating Algerian Arabic written in Arabizi and Arabic script into MSA, while~\newcite{baniata2018neural} introduce a system that translates Levantine (Jordanian, Syrian, Palestinian) and Maghrebi (Algerian, Moroccan, Tunisian) into MSA.~\newcite{sajjad2020arabench} propose an evaluation benchmark for Dialectal Arabic to English MT, along with several NMT systems using different training setups such as fine-tuning, data augmentation, and back-translation.~\newcite{farhan2020unsupervised} offer an unsupervised dialectal system where the source dialect (zero-shot) is not represented in training data.  \newcite{nagoudi-2021-Code-Mixed} propose a transformer-based MT system for translating from code-mixed MSA and Egyptian Arabic into English.
More recently, \newcite{kadaoui2023tarjamat} present a comprehensive evaluation of large language models (LLMs), including Bard and ChatGPT, on the machine translation of ten Arabic varieties.
To the best of our knowledge, our work is the first shared task to enable investigating MT in $four$ Arabic dialects, namely  \textit{Egyptian}, \textit{Emirati}, \textit{Jordanian}, and \textit{Palestinian}. For our MT subtasks, we also annotate and release a novel dataset and facilitate comparisons in a standardized experimental setting.

\subsection{Previous NADI Shared Tasks}

\paragraph{NADI 2020} The first NADI shared task,~\cite{mageed:2020:nadi} was co-located with the fifth Arabic Natural Language Processing Workshop (WANLP 2020)~\cite{wanlp-2020-arabic}. NADI 2020 targeted both country- and province-level dialects. It covered a total of $100$ provinces from $21$ Arab countries, with data collected from Twitter. It was the first shared task to target naturally occurring fine-grained dialectal text at the sub-country level. 

\paragraph{NADI 2021} The second edition of the shared task \cite{mageed:2021:nadi} was co-located with WANLP 2021~\cite{wanlp-2021-arabic}. It targeted the same $21$ Arab countries and $100$ corresponding provinces as NADI 2020, also exploiting Twitter data. NADI 2021 improved over NADI 2020 in that non-Arabic data were removed. In addition, NADI-2021 teased apart the data into MSA and DA and focused on classifying MSA and DA tweets into the countries and provinces from which they are collected. As such, NADI 2021 had four subtasks: MSA-country, DA-country, MSA-province, and DA-province. 

\paragraph{NADI 2022} The third edition of the shared task \cite{mageed:2022:nadi} was co-located with WANLP 2021.\footnote{\url{https://sites.google.com/view/wanlp2021}} It focused on studying Arabic dialects at the country level as well as dialectal sentiment (i.e., sentiment analysis of data tagged with dialect labels). 
We discuss NADI 2023 in more detail in the next section.

%% file: tasks.tex
\input{tables/examples_dia}
\input{tables/examples_mt}
\section{Task Description}\label{sec:tasks}
In NADI-2023, we place our emphasis on two NLP tasks, both crucial to processing of dialectal Arabic. Dialect identification remains an important step in any pipeline for processing dialects, for which reason NADI-2023 \textbf{Subtask~1} maintains the focus on identification of Arabic dialects. In particular, Subtask~1 targets dialect at the country level. Another important NLP task that has not particularly witnessed accelerated progress over the past few years is machine translation of Arabic dialects. For this reason, we take as our second focal point MT of dialects through \textbf{Subtask~2} and \textbf{Subtask~3}. We now describe each subtask in detail.

\subsection{Subtask 1: Dialect Identification }\label{subsec:tasks-DI}

Dialect identification has consistently been central to the NADI shared task over the years (\citeyear{mageed:2020:nadi,mageed:2021:nadi,mageed:2022:nadi}). In NADI-2023, we continue to focus on dialect identification through Subtask~1. 

\paragraph{Data} For this purpose, we provide a new Twitter dataset (i.e., TWT-2023), encompassing $18$ distinct dialects, totaling $23.4$K tweets. 
We also provide access to additional datasets for training. These are NADI-$2020$~\cite{mageed:2020:nadi}, NADI-$2021$~\cite{mageed:2021:nadi}, and MADAR~\cite{Bouamor:2018:madar} \textit{training} splits. We refer to these datasets as NADI-2020-TWT, NADI-2021-TWT, and MADAR-2018, respectively. We provide further details about these datasets in Section~\ref{subsec:dataset-eval-data}. Table~\ref{tab:examples_dia} shows examples from tweets in our NADI-2023 dataset for five countries. 

\paragraph{Restrictions} It is essential to note that Subtask~1 operates under a \textbf{\textit{closed-track}} policy where participants are allowed to use for system training \textit{only} datasets we provide. That is, no external data sources can be used for training purposes in this subtask.

\subsection{Subtasks 2 and 3: Machine Translation}\label{subsec:tasks-MT}
In this competition version, we introduce a new theme to NADI centered around machine translation from \textit{four} Arabic dialects to Modern Standard Arabic (MSA) at the sentence level. We present two versions of this competition, one is a closed track (Subtask~2), and the other is an open track (Subtask~3).

\paragraph{Dev and Test Data} For both Subtask~2 and Subtask~3, we manually curate new development and test datasets that each cover \textit{four} Arabic dialects: \textit{Egyptian}, \textit{Emirati}, \textit{Jordanian}, and \textit{Palestinian}. We refer to these new datasets as MT-2023-DEV and MT-2023-TEST, respectively. MT-2023-DEV comprises $400$ sentences, with $100$ sentences representing each of the four dialects; whereas MT-2023-TEST has a total of $2,000$ sentences, $500$ from each dialect. Table~\ref{tab:examples_mt} shows example sentences from MT-2023-DEV for each of the four countries. 
During the competition, we intentionally kept the source domain of these datasets undisclosed. Since we typically keep a live leaderboard for post-competition evaluation, we will not disclose the MT-2023* data domain.  

\paragraph{Restrictions} For the MT theme, restrictions on use of training datasets depend on the type of track. We offer two tracks, one closed and another open each with its own subtask. We introduce these subtasks now, detailing respective track information.

\paragraph{Subtask~2 -- Closed-Track Dialect to MSA MT} For Subtask~2 training, we restrict to the MADAR parallel dataset~\cite{bouamor2019madar}. More precisely, participants were allowed to use only the training split of MADAR parallel corpus for this subtask, and report on the development and test sets we provide. This meant that use of MADAR development and test datasets was not allowed for Subtask~2.

\paragraph{Subtask~3 -- Open-Track Dialect to MSA MT} For Subtask~3 training, participants were allowed to train their systems on any additional datasets of their choice so long as these additional training datasets are public at the time of submission. For example, participants were allowed to manually create new parallel datasets. For transparency and wider community benefits, we required researchers participating in the open track subtask to submit the datasets they create along with their Test set submissions. 

%% file: tables/examples_dia.tex
\begin{table*}[!ht]
\centering
 \renewcommand{\arraystretch}{1.25}
\resizebox{0.9\textwidth}{!}{%
\begin{tabular}{lr}
\toprule
\textbf{Country} & \multicolumn{1}{c}{\textbf{Content}} \\
\midrule
\multirow{2}{*}{Algeria} & \RL{ مهم جميع واحد انا ثقيل عليه يبلوكيني و ميضليش ينفخلي فيهم بهدره تع تقيي } \\
 & \RL{ يكونو معاك هاك يشوفوك وليت هاك يولو هاك و يكتلوك بهاك سورتو لا صابوك هاك يولو عليك غي هاك هوما هاك } \\ \midrule
\multirow{2}{*}{Iraq}  & \RL{ بس ما اعتقد لان هيج وقت كل الاكلات حلوه } \\
 & \RL{ بو خوشي بهس تهت فيت زيانه ته سبي بيت اوب ههمي دلي هه } \\\midrule
\multirow{2}{*}{Jordan}  & \RL{ من لما صحيت حاسه في اشي غريب اشي ناقص لحتي امي سالتني شربتي قهوه اليوم ها ها ها } \\
 & \RL{ هه ممكن والله كل اشي بهالبلد ممكن يصير } \\\midrule
\multirow{2}{*}{Saudi Arabia }  & \RL{ ذيك الايام خشم بخاري بغيبوبه } \\
& \RL{ وش اسمه يمدي حرمته تدعي علي هه } \\\midrule
\multirow{2}{*}{Sudan}  & \RL{ اكتر من 10 سنين شغاله يا استنكرت موضوع يا ذرفت دموع اسه جابت ليها قطع رؤوس \#محن\_الكيزان في \#السودان } \\
 & \RL{ اي زول في العلاقه عاوز اعمل تقيل بقتل العلاقه والله يعني الطرفين ساكتين لو في طرف م عاوز التاني اصرفها ليهو } \\
\bottomrule
\end{tabular}%
}
\caption{Random examples from NADI-2023 Subtask-1 training dataset spanning five different countries.}
\label{tab:examples_dia}
\end{table*}

%% file: tables/examples_mt.tex
\begin{table*}[!ht]
 \renewcommand{\arraystretch}{1.3}
\resizebox{\textwidth}{!}{%
\centering
\begin{tabular}{lrr}
\toprule
\textbf{Dialect } & \multicolumn{1}{r}{\textbf{Source (Dialect)~~~~~~~~~~~~~~~~~~~~~~~~~~~~~~~~~~~~~~~}} & \multicolumn{1}{r}{\textbf{Target (MSA)~~~~~~~~~~~~~~~~~~~~~~~~~~~~~~~~~~~~~~~}} \\
\bottomrule
\multicolumn{1}{c}{\multirow{3}{*}{\rotatebox[origin=c]{90}{Egyptain}} } & \RL{ ايوا حضرتك ده حق ابويا، حقي أنا بقى؟ } & \RL{ إي حضرتك هذا حق أبي ، أين حقي أنا إذن ؟ } \\
\multicolumn{1}{c}{}  & \RL{ حتى ابوك نفسه مش حيقدر يشفعلك عنده. } & \RL{ وأبوك نفسه لن يقدر يشفع لك عنده . } \\
\multicolumn{1}{c}{}  & \RL{ يا عم احنا مش ناقصينك الله يسترك عيب كده اختشي فوق فوق. } & \RL{ يا عم ! نحن فاض بنا الكيل ، الله يسترك ، هذا عيب ! استحِ ! أفِق ! أفِق! } \\ \midrule
\multicolumn{1}{c}{\multirow{3}{*}{\rotatebox[origin=c]{90}{Emirati}}}  & \RL{ زين خبروني شو السالفة ؟ } & \RL{ إذا أخبروني، ما القصة؟ } \\
\multicolumn{1}{c}{}  & \RL{ مابا حد يدري في الفريج، إن بو محمد انسرق } & \RL{ لا أريد أن يعلم أحدًا في الحي، أن أبو محمد انسرق } \\
\multicolumn{1}{c}{}  & \RL{ انزين، بغينا اثنين زنجبيل حار } & \RL{ حسنا، نريد اثنين من الزنجبيل الحار } \\ \midrule
\multicolumn{1}{c}{\multirow{3}{*}{\rotatebox[origin=c]{90}{Jordanian}} }  & \RL{ بس ما بعرف شو، مش عارف شو صارلي } & \RL{ لكن لا أعلم ماذا، لا أعلم ما حدث لي } \\
\multicolumn{1}{c}{}  & \RL{ كله منك انت السبب ليش ماخليتني ماسك بخناقه } & \RL{ كله بسببك أنت السبب لماذا لم تدعني ممسكاً بعنقه } \\
\multicolumn{1}{c}{}  & \RL{ بكير من عمرك } & \RL{ بارك الله في عمرك } \\\midrule
\multicolumn{1}{c}{\multirow{3}{*}{\rotatebox[origin=c]{90}{Palestinian}}}  & \RL{ طيب و هاي الملاعب وين؟ } & \RL{ حسناً، وهذه الملاعب أين؟ } \\
 \multicolumn{1}{c}{} & \RL{ شو يا أبو ناصر سمعت صاير تهدد في القتل } & \RL{ ما هذا يا أبو ناصر، سمعت أنك أصبحت تهدد بالقتل } \\
 \multicolumn{1}{c}{} & \RL{ شاطر، إلك عندي باكيت حلقوم } & \RL{ حصيف، سأكافئك بعلبة كاملة من حلوى الحلقوم} \\ \bottomrule
\end{tabular}%
}
\caption{Random examples from MT-2023-DEV dataset spanning the four covered dialects.}
\label{tab:examples_mt}
\end{table*}

%% file: datasets.tex
\input{tables/data}
\section{Shared Task Datasets and Evaluation}\label{sec:dataset-eval}

In this section, we describe the datasets we make available to participants, introduce the chosen evaluation metrics, and outline the clear instructions we provided for the submission process.

\subsection{Datasets}\label{subsec:dataset-eval-data}
\begin{itemize} 
    \item \textbf{TWT-2023}: \newcite{mageed2020microdialect} introduce a vast dataset comprising $\sim$$6$B tweets from $2.7$M users. They systematically extract tweets that contain geographic information and subsequently embark on a manual annotation process for each user, classifying their location at the city, state, and country levels. This effort results in the identification of $\sim500$M tweets coming from $233$K users spread across $319$ cities within $21$ Arab countries. For Subtask~1, we randomly select from this data $1,000$ training, $100$ development, and $200$ testing tweets for each of the $18$ covered countries. In total, this amounts to $23,400$ tweets that we refer to as TWT-2023. We split TWT-2023 into Train ($18$K), Dev ($1.8$K), and Test ($3.6$K). 
    \item \textbf{NADI-202X-TWT}. We also distribute \textbf{NADI-2020-TWT} and \textbf{NADI-2021-TWT} datasets. These datasets are similarly collected from Twitter. For both of them, we use the Twitter API to crawl data from $21$ Arab countries for a period of $10$ months (Jan. to Oct., $2019$). For each case, we label tweets from each user with the country from which they posted for the whole of the $10$ months period, thus exploiting \textit{consistent posting location} as a proxy for \textit{dialect labels}. We use the same training splits as both NADI-2020 and NADI-2021, but only include data that cover the $18$ Arab countries we target in the current 2023 edition. It is also noteworthy that we do not provide the NADI-2022 training dataset since it is identical to the training set used in NADI 2021.
    \item \textbf{MADAR-18}: The MADAR corpus is a collection of parallel sentences encompassing the dialects of $25$ cities from across the Arab world, along with English, French, and MSA. Since this dataset does not originally have country-level labels, we  map the $25$ cities to their respective countries. As a result, we acquire a customized version of MADAR that we refer to as MADAR-18. We offer the dialectal side of MADAR-18 for optional use for training systems for Subtask-1.
    \item \textbf{MADAR-4-MT}: We extract parallel dialectal-to-MSA data of four dialects from MADAR-18 for training MT systems for Subtask-2 and Subtask-3. The four pairs involve Egyptian, Emirati, Jordanian, and Palestinian at the dialectal side.

\end{itemize}
Table~\ref{tab:subtask1_data} present the statistics and characteristics of NADI-2023's Subtask-1 training, development, and test datasets, along with the distribution of our additional resources, i.e, NADI-2020-TWT, NADI-2021-TWT, and MADAR-18.\footnote{Recall that MADAR-4-MT is extracted from MADAR-2018}

\subsection{Evaluation Metrics}\label{subsec:dataset-eval-metrics}

The official evaluation metric for Subtask-1 is the macro-averaged \texttt{F\textsubscript{1}} score. In addition to this metric, we also report system performance in terms of  \texttt{Precision}, \texttt{Recall}, and \texttt{Accuracy} for submissions to this Subtask~1. For both Subtask~2 and Subtask~3, we use the \texttt{Bleu} score as the official metric. The \texttt{Bleu} score is computed separately for each of the four dialects (i.e., Egyptian, Emirati, Jordanian, and Palestinian). We then use the average of these individual Bleu scores to rank the submitted systems for Subtask~2 and Subtask~3.

\subsection{Submission Roles}\label{subsec:dataset-eval-roles}
We allowed participant teams to submit up to \textit{five} runs for each test set, for each of the three subtasks. In each case, we only retain the submission with the highest score for each team. While the official results were exclusively based on a blind test set, we also requested participants to include their results on the development datasets (Dev) in their papers. 

To facilitate the evaluation of participant systems, we established a CodaLab competition for scoring each subtask (i.e., a total of three Codalabs).\footnote{The different CodaLab competitions are available at the following links: \href{https://codalab.lisn.upsaclay.fr/competitions/14449}{\texttt{subtask-1}}, \href{https://codalab.lisn.upsaclay.fr/competitions/14643}{\texttt{subtask-2}}, and \href{https://codalab.lisn.upsaclay.fr/competitions/14648}{\texttt{subtask-3}}.} Similar to previous NADI editions, we are keeping the CodaLab for each subtask active even after official competition has concluded. This is to encourage researchers interested in training models and assessing systems using the shared task's blind test sets. Consequently, we will not disclose the labels for the test sets of any of the subtasks.

%% file: tables/data.tex
\begin{table}[t]
\centering
 \renewcommand{\arraystretch}{1.25}
\resizebox{0.47\textwidth}{!}{%
\begin{tabular}{lHHHrrrr}
\toprule
\textbf{Country} & \textbf{TRAIN} & \textbf{DEV} & \textbf{TEST} & \multirow{-2}{*}{} & \textbf{NADI-2020} & \textbf{NADI-2021} & \textbf{MADAR-18} \\ \midrule

Algeria & $1,000$ & $100$ & $200$ &  & $1,491$ & $1,809$ & $1,600$ \\
Bahrain & $1,000$ & $100$ & $200$ &  & $210$ & $215$ & $-~~~~$ \\
Egypt & $1,000$ & $100$ & $200$ &  & $4,473$ & $4,283$ & $4,800$ \\
Iraq & $1,000$ & $100$ & $200$ &  & $2,556$ & $2,729$ & $4,800$ \\
Jordan & $1,000$ & $100$ & $200$ &  & $426$ & $429$ & $3,200$ \\
Kuwait & $1,000$ & $100$ & $200$ &  & $420$ & $429$ & $-~~~~$ \\
Lebanon & $1,000$ & $100$ & $200$ &  & $639$ & $644$ & $1,600$ \\
Libya & $1,000$ & $100$ & $200$ &   & $ 1,070  $ & $ 1,286  $ & $ 3,200$ \\
Morocco & $1,000$ & $100$ & $200$ &   & $ 1,070  $ & $ 858  $ & $ 3,200$ \\
Oman & $1,000$ & $100$ & $200$ &   & $ 1,098  $ & $ 1,501  $ & $ 1,600$ \\
Palestine & $1,000$ & $100$ & $200$ &   & $ 420  $ & $ 428  $ & $ 1,600$ \\
Qatar & $1,000$ & $100$ & $200$ &   & $ 234  $ & $ 215  $ & $ 1,600$ \\
Saudi Arabia & $1,000$ & $100$ & $200$ &   & $ 2,312  $ & $ 2,140  $ & $ 3,200$ \\
Sudan & $1,000$ & $100$ & $200$ &   & $ 210  $ & $ 215  $ & $ 1,600$ \\
Syria & $1,000$ & $100$ & $200$ &   & $ 1,070  $ & $ 1,287  $ & $ 3,200$ \\
Tunisia & $1,000$ & $100$ & $200$ &   & $ 750  $ & $ 859  $ & $ 3,200$ \\
UAE & $1,000$ & $100$ & $200$ &   & $ 1,070  $ & $ 642  $ & $ -~~~~$ \\
Yemen & $1,000$ & $100$ & $200$ &   & $ 851  $ & $ 429  $ & $ 1,600$ \\ \midrule
\textbf{Total} & $\bf 18,000$ & $\bf 1,800$ & $\bf 3,600$ & & $\bf 20,370$ & $\bf 20,398$ & $\bf 40,000$  \\ \bottomrule
\end{tabular}%
}
\caption{Distribution of Subtask-1 additional training data. For NADI-2023, we also distribute a total of $18,000$ tweets for Train, $1,800$ for Dev, and $3,600$ for Test (with $1,000$, $100$, and $200$ from each country for $18$ countries listed in the table for Train, Dev, and Test, respectively). For Subtask 2 and Subtask-3, we extract MADAR-4-MT from Egyptian, Emirati, Jordanian, and Palestinian data in MADAR-18 (see Section~\ref{sec:dataset-eval}). }
\label{tab:subtask1_data}
\end{table}


%% file: results.tex
\section{Shared Task Teams \& Results}\label{sec:results}

\input{tables/registered_team}
\input{tables/subtask1_res}
\input{tables/subtask2_res}
\input{tables/subtask3_res}

\input{tables/sum_methods}
\input{tables/sum_methods_sub_task3}
\subsection{Participating Teams}
\label{sec:teams}

We received a total of $58$ unique team registrations. At the testing phase,  a total of $76$ valid entries were submitted by $18$ unique teams. The breakdown across the subtasks is as follows: $49$ submission for Subtask~1 from $16$ teams, $16$ submissions for Subtask~2 from $5$ teams, and $11$ submissions for Subtask~3 from $3$ teams. Table~\ref{tab:teams} lists the $18$ teams. A total of $14$ teams submitted $14$ description papers from which we accepted $13$ papers for publication. Accepted papers are cited in Table~\ref{tab:teams}.

\subsection{Baselines}\label{subsec:results-baselines}

For comparison,  we provide three baselines for each of the three subtasks. For \textbf{Subtask 1}, we finetune MARBERT\textsubscript{v2}~\cite{mageed2020marbert},  AraBERT\textsubscript{twitter}~\cite{araelectra-2021-antoun}, and CAMeLBERT\textsubscript{da}~\cite{obeid-etal-2020-camel},  on \texttt{TWT-2023} training data (see Section~\ref{subsec:tasks-DI}).  
For \textbf{Subtask 2} and \textbf{3},\footnote{We use the same baseline models to both Subtask 2 and Subtask 3.}  we finetune AraT5\textsubscript{v2}~\cite{nagoudi-2022-arat5}, mT5~\cite{xue2020mt5}, and AraBART~\cite{eddine2022arabart} on MADAR-4-MT (see Section~\ref{subsec:tasks-DI}). In each subtask, we label these baselines as \textbf{Baseline I}, \textbf{II}, and \textbf{III}, respectively.

For all the baselines in both tasks, we finetune each model using the training data specific to each subtask (i.e., TWT-2023 for Subtask~1 and MADAR-4-MT for Subtask~2 and Subtask~3) for $10$ epochs with a learning rate of $2e-5$ and batch size of $32$. The maximum length is set to $256$ tokens and we set an early stopping patience to $5$. Following each epoch, we evaluate each model and select the best the best-performing model on the respective Dev set. Subsequently, we present the performance metrics of this best-performing model on the test datasets.


\subsection{Shared Task Results}\label{subsec:results-discussion}
Table~\ref{tab:subtask1_res} presents the leaderboard of Subtask~1 and is sorted by macro-$F_1$. As Table~\ref{tab:subtask1_res} shows, for each team, we take their best macro-$F_1$ score to represent them. \texttt{Team NLPeople}~\cite{nlpeople-2023-nadi} obtained the best performance on Subtask~1 with $87.27$ macro-$F_1$. We can observe that $9$ teams outperform our strongest baseline, MARBAET (i.e, Baseline I). 
Table~\ref{tab:subtask2_res} and Table~\ref{tab:subtask3_res} show the leaderboard of Subtask~2 and 3, respectively. Both are sorted by their main metrics, the overall BLEU score. \texttt{Team UniManc}~\cite{unimanc-2023-nadi} won both subtasks, achieving the best BLEU scores of $14.76$ and $21.10$ on Subtask~2 and 3, respectively. We observe that \textit{five} teams outperform our Baseline~I on Subtask~2.

\subsection{General Description of Submitted Systems}\label{subsec:results-papers-desc}

In Tables~\ref{tab:system_sum_subtask1} and \ref{tab:system_sum_subtask23}, we provide a high-level summary of the submitted systems. For each team, we list the best score with the main metric of each subtask and the number of submissions made by the team. As shown in these tables, most teams use pretrained language models (PLM), including Transformer encoder-based PLMs (e.g., AraBERT~\cite{antoun2020arabert} and MARBERT~\cite{mageed2020marbert}) for Subtask~1 and Transformer encoder-decoder PLMs (e.g., ArabT5~\cite{nagoudi-2022-arat5}) for Subtask~2 and Subtask~3. Ensemble voting is also an effective approach most teams employ in Subtask~1. 

The top team of Subtask~1, i.e., \texttt{NLPeople}~\cite{nlpeople-2023-nadi}, exploits MARBERT, AraBERT, and AraT5 with different finetuning strategies (e.g., staged finetuning). To enrich the learning context, they use a retrieval method to find similar texts from the training set for a given text and then append the retrieved texts along with corresponding labels as additional input. Their best submission is an ensemble with ten models.
Team \texttt{rematchka}~\cite{rematchka-2023-nadi}, exploits MARBERT, AraBERT, AraELECTRA~\cite{araelectra-2021-antoun}, and CAMeLBERT~\cite{obeid-etal-2020-camel} with different prompting techniques and add linguistic features to their models. They also use supervised contrastive loss~\cite{supervised-2021-gunel} to enhance model finetuning. Teams \texttt{SANA}~\cite{sana-2023-nadi} and \texttt{Frank}~\cite{frank-2023-nadi} both finetune PLMs and apply ensemble voting to achieve their best performance. 

On Subtask~2 (closed track), the winning team, \texttt{Team UniManc}~\cite{unimanc-2023-nadi}, finetune three variants of T5 models (i.e., mT5~\cite{xue-etal-2021-mt5}, mT0~\cite{crosslingual-2023-muennighoff}, and AraT5) with the officially released dataset. For Subtask~3 (open track), \texttt{Team UniManc} collects four additional supervised datasets and uses GPT-3.5-turbo to translate $2,712$ samples from Subtask~1. Team \texttt{Helsinki-NLP}~\cite{helsinki-2023-nadi} finetune ByT5~\cite{xue-etal-2022-byt5} and AraT5 with the officially released dataset of Subtask~2. For Subtask~3, they collect six monolingual MSA datasets and synthesize a parallel dataset by exploiting character-level statistical machine translation models to translate the MSA to different dialects. They then finetune PLMs with the supervised dataset from Subtask~2 and their synthetic dataset. Similarly, both teams \texttt{DialectNLU} and \texttt{rematchka} finetune AraT5 with the training data of Subtask~2.

%% file: tables/registered_team.tex
\begin{table*}[!ht]
\centering
 \renewcommand{\arraystretch}{1.25}
\resizebox{0.75\textwidth}{!}{%
\begin{tabular}{llc}
\toprule
\textbf{Team} & \textbf{Affiliation}                           & \textbf{Tasks} \\ \midrule
AIC                              & Applied Innovation Center, Egypt               & $1$                                  \\
ANLP-RG~\cite{anlp-rg-2023-nadi}                               & University of Sfax, Tunisia               & $2$                                  \\
Arabitools                        & STEAM Solutions, Palestine                     & $1$                                  \\
Cordyceps                         & University of Toronto, Canada                  & $1$                                  \\
DialectNLU~\cite{dialectnlu-2023-nadi}                        & UCLA, USA                                      & $1$, $2$                               \\
Exa                              & Exa, Iran                                      & $1$                                  \\
Frank~\cite{frank-2023-nadi}                             & MBZUAI, UAE                                    & $1$                                  \\
Fraunhofer IAIS                 & Fraunhofer IAIS, Germany                       & $1$, $2$                               \\
Helsinki-NLP~\cite{helsinki-2023-nadi}                      & University of Helsinki, Finland                & $2$, $3$                               \\
ISL-AAST~\cite{isl-aast-2023-nadi}                          & Arab Academy for Science and Technology, Egypt & $1$                                  \\
IUNADI~\cite{iunadi-2023-nadi}                            & Indiana University Bloomington, USA            & $1$                                  \\
Mavericks~\cite{mavericks-2023-nadi}                         & Pune Institute of Computer Technology, India   & $1$                                  \\
NAYEL                             & Benha University, Egypt                        & $1$                                  \\
NLPeople~\cite{nlpeople-2023-nadi}                          & IBM Research Europe, UK                        & $1$                                  \\
rematchka~\cite{rematchka-2023-nadi}                         & Cairo University, Egypt                        & $1$, $2$, $3$                            \\
SANA~\cite{sana-2023-nadi}                              & Taibah University, KSA                         & $1$                                  \\
UniManc~\cite{unimanc-2023-nadi}                           & University of Manchester, UK                   & $2$, $3$                               \\
UoT~\cite{uot-2023-nadi}                               & University of Tripoli, Libya                   & $1$                                  \\
usthb~\cite{usthb-2023-nadi}                             & USTHB, Alegria                                 & $1$                                  \\ \bottomrule
\end{tabular}%
}
\caption{List of teams that participated in NADI-2023 shared task. Teams with accepted papers are cited.}\label{tab:teams}

\end{table*}

%% file: tables/subtask1_res.tex
\begin{table}[!ht]
\centering
 \renewcommand{\arraystretch}{1.3}
\resizebox{0.48\textwidth}{!}{%
\begin{tabular}{@{}clrrrr@{}}
\toprule
\textbf{Rank} & \textbf{Team}   & \textbf{F1} & \textbf{Acc.} & \textbf{Pre.} & \textbf{Rec.} \\ \midrule
1             & NLPeople        & 87.27       & 87.22             & 87.37              & 87.22           \\
2             & rematchka       & 86.18       & 86.17             & 86.29              & 86.17           \\
3             & Arabitools      & 85.86       & 85.81             & 86.10              & 85.81           \\
4             & SANA            & 85.43       & 85.39             & 85.60              & 85.39           \\
5             & Frank           & 84.76       & 84.75             & 84.95              & 84.75           \\
6             & ISL-AAST        & 83.73       & 83.67             & 83.87              & 83.67           \\
7             & UoT             & 82.87       & 82.86             & 83.17              & 82.86           \\
8             & AIC             & 82.37       & 82.42             & 82.57              & 82.42           \\
9             & Cordyceps       & 82.17       & 82.14             & 82.57              & 82.14           \\ 
\cdashline{1-6}
\multicolumn{1}{l}{\textcolor{blue}{Baseline I}}          &  {MARBERT\textsubscript{v2}}         & 81.44 & 81.36 & 81.68 & 81.36  \\ \cdashline{1-6}

10            & DialectNLU      & 80.56       & 80.50             & 80.92              & 80.50           \\ \cdashline{1-6}
\multicolumn{1}{l}{\textcolor{blue}{Baseline II}}          & {AraBERT\textsubscript{twitter}   }        & 77.02 & 76.97 & 77.54 & 76.97     \\ \cdashline{1-6}
11            & Mavericks       & 76.65       & 76.47             & 77.43              & 76.47           \\ \cdashline{1-6}
\multicolumn{1}{l}{\textcolor{blue}{Baseline III}}            & {CAMeLBERT\textsubscript{da}    }       &  74.56 & 74.47 & 74.90 & 74.47    \\ \cdashline{1-6}
12            & exa             & 70.72       & 71.03             & 72.26              & 71.03           \\

13            & IUNADI          & 70.22        & 70.78              & 71.32               & 70.78            \\

14            & NAYEL           & 63.09       & 63.39             & 63.30              & 63.39           \\
15            & usthb           & 62.51       & 62.17             & 63.07              & 62.17           \\
16            & Fraunhofer IAIS & 29.91       & 33.14             & 38.47              & 31.39           \\

\bottomrule
\end{tabular}%
}
\caption{Results of Subtask 1 (Country-Level DA).}\label{tab:subtask1_res}
\end{table}

%% file: tables/subtask2_res.tex
\begin{table}[]
\centering
 \renewcommand{\arraystretch}{1.3}
\resizebox{0.45\textwidth}{!}{%
\begin{tabular}{@{}clrrrrr@{}}
\toprule
\textbf{Rk} & \textbf{Team}   & \textbf{Overall} & \textbf{Egy.} & \textbf{Emi.} & \textbf{Jor.} & \textbf{Pal.} \\ \midrule
1             & UniManc         & 14.76           & 16.04             & 14.30            & 12.55              & 13.55                \\
2             & Helsinki        & 14.28           & 12.22             & 23.13            & 11.15              & 13.42                \\
3             & DialectNLU      & 13.43           & 11.45             & 21.59            & 10.64              & 12.66                \\
4             & rematchka       & 11.37           & 11.18             & 11.99            & 10.47              & 10.86                \\
5             & ANLP-RG & 10.02            & 10.25              & 8.50             & 10.26              & 9.33                 \\ \cdashline{1-7}

\multicolumn{1}{l}{\textcolor{blue}{Baseline I}}            &   AraT5\textsubscript{v2}  &  7.70  &  5.50  &  10.45  &  9.51  &  6.48   \\\cdashline{1-7}

6             & Fraunhofer IAIS & 5.85            & 8.08              & 3.90             & 4.96               & 6.01                 \\ \cdashline{1-7}

\multicolumn{1}{l}{\textcolor{blue}{Baseline II}}            & mT5 & 2.98  &  4.17  &  3.66  &  3.89  &  3.95    \\

\multicolumn{1}{l}{\textcolor{blue}{Baseline III}}         & AraBART &2.63  &  2.44  &  3.16  &  1.89  &  2.60   \\
\bottomrule
\end{tabular}%
}
\caption{Results of Subtask 2 (Closed AD to MSA MT)}\label{tab:subtask2_res}
\end{table}

%% file: tables/subtask3_res.tex
\begin{table}[!ht]
\centering
 \renewcommand{\arraystretch}{1.3}
\resizebox{0.49\textwidth}{!}{%
\begin{tabular}{@{}clrrrrr@{}}
\toprule
\textbf{Rank} & \multicolumn{1}{c}{\textbf{Team}} & \multicolumn{1}{c}{\textbf{Overall}} & \multicolumn{1}{c}{\textbf{Egy.}} & \multicolumn{1}{c}{\textbf{Emi.}} & \multicolumn{1}{c}{\textbf{Jor.}} & \multicolumn{1}{c}{\textbf{Pal.}} \\ \midrule
1             & UniManc                           & 21.10                                & 17.65                             & 28.46                             & 22.03                             & 17.29                             \\
2             & Helsinki-NLP                      & 17.69                                & 16.11                             & 25.81                             & 15.60                             & 15.91                             \\
3             & rematchka                         & 11.37                                & 11.18                             & 11.99                             & 10.47                             & 10.86                             \\ \cdashline{1-7}

\multicolumn{1}{l}{\textcolor{blue}{Baseline I}}            &   AraT5\textsubscript{v2}  &  5.41  &  5.50  &  5.84  &  6.06  &  4.47    \\
\multicolumn{1}{l}{\textcolor{blue}{Baseline II}}             & mT5 & 2.98  &  4.17  &  3.66  &  3.89  &  3.95    \\

\multicolumn{1}{l}{\textcolor{blue}{Baseline III}}        & AraBART &1.12  &  0.00  &  0.00  &  1.17  &  1.10     \\
\bottomrule
\end{tabular} %
}
\caption{Results of Subtask 3 (Open DA to MSA MT).}\label{tab:subtask3_res}
\end{table}


%% file: tables/sum_methods.tex
\begin{table*}[!ht]
\centering
 \renewcommand{\arraystretch}{1.25}
\resizebox{0.95\textwidth}{!}{%
\begin{tabular}{@{}llllllllllllllll@{}}
\toprule
\multirow{2}{*}{\textbf{Team}} & \multicolumn{1}{c}{\multirow{2}{*}{\rotatebox[origin=c]{70}{\textbf{\# submit}}}} & \multicolumn{1}{c}{\multirow{2}{*}{{\textbf{F\textsubscript{1}}}}} & \multicolumn{5}{c}{\textbf{Features}}                                                                                                                                                                      & \multicolumn{8}{c}{\textbf{Techniques}}                                                                                                                                                                                                                                                                                         \\ \cmidrule(l){4-7} \cmidrule(l){8-16}
                                    & \multicolumn{1}{c}{}                                    & \multicolumn{1}{c}{}                                      & \multicolumn{1}{c}{\textbf{\rotatebox[origin=c]{70}{N-gram}}} & \multicolumn{1}{c}{\rotatebox[origin=c]{70}{\textbf{TF-IDF}}} & \multicolumn{1}{c}{\textbf{\rotatebox[origin=c]{70}{Linguistic}}} & \multicolumn{1}{c}{\rotatebox[origin=c]{70}{\textbf{Word embeds}}} & \multicolumn{1}{c}{\textbf{\rotatebox[origin=c]{70}{Classical ML}}} & \multicolumn{1}{c}{\rotatebox[origin=c]{70}{\textbf{Neu. nets}}} & \multicolumn{1}{c}{\rotatebox[origin=c]{70}{\textbf{PLM}}} & \multicolumn{1}{c}{\rotatebox[origin=c]{70}{\textbf{Ensemble}}} & \multicolumn{1}{c}{\rotatebox[origin=c]{70}{\textbf{Adapter}}} & \multicolumn{1}{c}{\rotatebox[origin=c]{70}{\textbf{Hie. Cls}}} & \multicolumn{1}{c}{\rotatebox[origin=c]{70}{\textbf{Prompting}}} & \multicolumn{1}{c}{\rotatebox[origin=c]{70}{\textbf{Contrast. L}}} & \multicolumn{1}{c}{\rotatebox[origin=c]{70}{\textbf{Data Aug.}}} \\ \midrule
\textbf{NLPeople}                   & 5                                                       & 87.27                                                     &                                     &                                     &                                         &                                          & \checkmark                                      & \checkmark                                     & \checkmark                             & \checkmark                                  &                                      &                                       &                                        &                                          & \checkmark                                   \\
\textbf{rematchka}                  & 3                                                       & 86.18                                                     &                                     &                                     &                                         &                                          &                                           &                                          & \checkmark                             & \checkmark                                  & \checkmark                                 &                                       & \checkmark                                   & \checkmark                                     &                                        \\
\textbf{Arabitools}                 & 4                                                       & 85.86                                                     &                                     &                                     & \checkmark                                    &                                          &                                           &                                          & \checkmark                             & \checkmark                                  &                                      &                                       &                                        &                                          &                                        \\
\textbf{SANA}                       & 2                                                       & 85.43                                                     &                                     &                                     &                                         &                                          &                                           &                                          & \checkmark                             & \checkmark                                  &                                      &                                       &                                        &                                          &                                        \\
\textbf{Frank}                      & 2                                                       & 84.76                                                     &                                     &                                     &                                         &                                          &                                           &                                          & \checkmark                             & \checkmark                                  &                                      &                                       &                                        &                                          &                                        \\
\textbf{ISL-AAST}                   & 5                                                       & 83.73                                                     &                                     &                                     &                                         &                                          &                                           & \checkmark                                     & \checkmark                             & \checkmark                                  &                                      &                                       &                                        &                                          &                                        \\
\textbf{UoT}                        & 2                                                       & 82.87                                                     &                                     &                                     &                                         &                                          & \checkmark                                      &                                          & \checkmark                             &                                       &                                      &                                       &                                        &                                          & \checkmark                                   \\
\textbf{AIC}                        & 5                                                       & 82.37                                                     & \checkmark                                &                                     & \checkmark                                    &                                          &                                           &                                          & \checkmark                             & \checkmark                                  &                                      & \checkmark                                  &                                        &                                          & \checkmark                                   \\
\textbf{Cordyceps}                  & 4                                                       & 82.17                                                     &                                     &                                     &                                         &                                          &                                           &                                          &                                  &                                       &                                      &                                       &                                        &                                          &                                        \\
\textbf{DialectNLU}                 & 5                                                       & 80.56                                                     &                                     & \checkmark                                & \checkmark                                    &                                          &                                           & \checkmark                                     & \checkmark                             & \checkmark                                  &                                      &                                       &                                        &                                          &                                        \\
\textbf{IUNADI}                     & 1                                                       & 70.22                                                     &                                     &                                     & \checkmark                                    &                                          &                                           &                                          & \checkmark                             & \checkmark                                  &                                      &                                       &                                        &                                          &                                        \\ 
\textbf{Mavericks}                  & 1                                                       & 76.65                                                     &                                     &                                     &                                         &                                          &                                           &                                          & \checkmark                             & \checkmark                                  &                                      &                                       &                                        &                                          &                                        \\
\textbf{NAYEL}                      & 5                                                       & 63.09                                                    & \checkmark                                & \checkmark                                &                                     &                                     & \checkmark                                      &                                          &                                  &                                       &                                      &                                       &                                        &                                          &                                  \\
\textbf{usthb}                      & 3                                                       & 62.51                                                     & \checkmark                                & \checkmark                                & \checkmark                                    & \checkmark                                     & \checkmark                                      &                                          &                                  &                                       &                                      &                                       &                                        &                                          & \checkmark                                   \\

\bottomrule
\end{tabular}%
}

\caption{Summary of approaches used by participating teams in Subtask 1. Teams are sorted by their performance on the official metric, Macro-$F_1$ score. Classical machine learning (ML) indicates any non-neural machine learning methods such as naive Bayes and support vector machines. The term ``neural nets" refers to any model based on neural networks (e.g., FFNN, RNN, CNN, and Transformer) trained from scratch. PLM refers to neural networks pretrained with unlabeled data such as MARBERT. (Hie. Cls, hierarchical classification approach); (Contrast. L, contrastive learning); (Data Aug., data Augmentation).}
\label{tab:system_sum_subtask1}
\end{table*}


%% file: tables/sum_methods_sub_task3.tex
\begin{table}[!ht]
\centering
\centering
 \renewcommand{\arraystretch}{1.3}
\setlength{\tabcolsep}{4pt}
\resizebox{0.47\textwidth}{!}{%
\begin{tabular}{lccccccc}
\toprule

\multirow{2}{*}{\textbf{Team}}& \multirow{2}{*}{\textbf{\# submit}}  & \multirow{2}{*}{\textbf{BLUE}} 
 \textbf{} & \multicolumn{5}{c}{\textbf{Techniques}} \\ 

\cmidrule(l){4-8}

& & & \textbf{Classic ML} & \textbf{NN} & \textbf{PLM} & \textbf{Ensemble} &{\textbf{Aug.}} \\ \midrule


&&&\multicolumn{5}{c}{\textbf{Subtask 2}}    \\ \midrule
\textbf{UniManc}  & 5     & 14.76    &     &    &   \checkmark  & \checkmark &   \\
\textbf{Helsinki} & 3     & 14.28   &    \checkmark  & \checkmark   &  \checkmark   &  &   \\
\textbf{DialectNLU}     & 5     & 13.43   &     &   &   \checkmark   &  \checkmark  &   \\
\textbf{rematchka}& 1     & 11.37   &     &  \checkmark   &   \checkmark   &  &   \\
\textbf{ANLP-RG}& 1     & 10.02   &  &      \checkmark &    \checkmark &    &    \checkmark   \\
\midrule
&&&\multicolumn{5}{c}{\textbf{Subtask 3}}   \\ \midrule
\textbf{UniManc}  & 5     & 21.10   &     &     & \checkmark   & \checkmark & \checkmark  \\
\textbf{Helsinki-NLP}   & 5     & 17.69  &    \checkmark & \checkmark  &  \checkmark  &  &  \checkmark \\
\textbf{rematchka}& 1     & 11.37   &    &   &   \checkmark &  &   \\

\bottomrule
\end{tabular}
}
\caption{Summary of approaches used by participating teams in Subtask 2 and 3. Teams are sorted by their performance on BLEU score for both Subtasks. Classical machine learning (ML) indicates any non-neural machine learning methods such as naive Bayes and support vector machines. "NN" refers to any model based on neural networks (e.g., FFNN, RNN, CNN, and Transformer) trained from scratch. PLM refers to neural networks pretrained with unlabeled data such as AraT5. (Aug., data augmentation).}
\label{tab:system_sum_subtask23}
\end{table}

%% file: conc.tex
\section{Conclusion}\label{sec:conc}
We presented findings and results of NADI-2023, the fourth edition of the NADI shared task focused on fine-grained Arabic dialect identification. This edition also introduced two subtasks centered on machine translation from four Arabic dialects into MSA. Results acquired by participant teams show that dialect identification remains a challenging task but that various types of approaches, many of which involve exploiting language models, can be used to handle the task. Similarly, translating Arabic dialects is unsurprisingly very challenging due to lack of training data. In the future, we plan to continue supporting both dialect identification and machine translation through NADI. 

%% file: limitations_ethics.tex
\section{Limitations}\label{sec:limis}

Our work has a number of limitations, as follows:

\begin{itemize}
\item Although we strive for widest coverage, this edition of NADI focused on only $18$ country-level dialects. This is due to our inability to develop high quality datasets for a few countries such as \textit{Comoros},  \textit{Djibouti}, \textit{Mauritania}, and \textit{Somalia}.
\item NADI continues to use short texts for the Arabic dialects. Due to lack of dialectal data from other sources, we depend on short posts from Twitter. Although these data have thus far empowered development of effective dialect identification models, it is desirable to afford data from other domains that have longer texts. This will allow development of more widely applicable models. 
\item Our MADAR-18 dataset is commissioned and, although useful, should not be used to analyze Arabic dialects as a replacement for naturally occurring data. 
\item Our machine translation subtasks focus only on four dialects and do not offer sizeable datasets. Modern MT systems need much larger data to perform well. Again, in spite of our best efforts, parallel datasets involving dialects remain limited. 
\end{itemize}

\section{Ethical Considerations}\label{sec:ethics}
The NADI-2023 Subtask~1 dataset is sourced from the public domain (i.e., X former Twitter), with user personal information and identity carefully concealed. Similarly, the NADI-2023 Subtask~2 and Subtask~3 datasets are manually created. Again, we take meticulous measures to remove user identities and personal information from this dataset. As a result, we have minimal concerns about the retrieval of personal information from our data. However, it is crucial to acknowledge that the datasets we collect to construct NADI-2023 Subtask~1 may contain potentially harmful content. Additionally, during model evaluation, there is a possibility of exposure to biases that could unintentionally generate problematic content. 

%% file: ack.tex
\section*{Acknowledgments}\label{sec:acknow}
We acknowledge support from Canada Research Chairs (CRC), the Natural Sciences and Engineering Research Council of Canada (NSERC; RGPIN-2018-04267), the Social Sciences and Humanities Research Council of Canada (SSHRC; 435-2018-0576; 895-2020-1004; 895-2021-1008), Canadian Foundation for Innovation (CFI; 37771), Digital Research Alliance of Canada,\footnote{\href{https://alliancecan.ca}{https://alliancecan.ca}} and UBC Advanced Research Computing-Sockeye.\footnote{\href{https://arc.ubc.ca/ubc-arc-sockeye}{https://arc.ubc.ca/ubc-arc-sockeye}}